\documentclass{article}

\usepackage{microtype}
\usepackage{graphicx}
\usepackage{subcaption}
\usepackage{booktabs} % for professional tables

\usepackage{hyperref}

\usepackage[preprint]{icml2026}

\usepackage{amsmath}
\usepackage{amssymb}
\usepackage{mathtools}
\usepackage{amsthm}
\usepackage[dvipsnames]{xcolor}

\newcommand{\rev}[1]{#1}

\usepackage[capitalize,noabbrev]{cleveref}

\theoremstyle{plain}

\theoremstyle{definition}

\theoremstyle{remark}

\usepackage[textsize=tiny]{todonotes}

\usepackage{siunitx}
\usepackage{makecell}

\icmltitlerunning{Not All Denoising Steps Are Equal: Model Scheduling for Faster Masked Diffusion Language Models}

\begin{document}

\twocolumn[
  \icmltitle{Not All Denoising Steps Are Equal: Model Scheduling for Faster Masked Diffusion Language Models}



  \icmlsetsymbol{equal}{*}

  \begin{icmlauthorlist}
    \icmlauthor{Ivan Sedykh}{mts}
    \icmlauthor{Nikita Sorokin}{mts}
    \icmlauthor{Valentin Malykh}{mts,uni}

  \end{icmlauthorlist}

  \icmlaffiliation{mts}{MWS AI}
  \icmlaffiliation{uni}{ITMO University}

  \icmlcorrespondingauthor{Ivan Sedykh}{i.sedykh@mts.ai}

  \icmlkeywords{Machine Learning, ICML}

  \vskip 0.3in
]



\printAffiliationsAndNotice{}  % no special notice (required even if empty)

% TITLE  
% Not All Denoising Steps Are Equal: Model Scheduling for Faster Masked Diffusion Language Models

\begin{abstract}
Recent advances in masked diffusion language models (MDLMs) narrow the quality gap to autoregressive LMs, but their sampling remains expensive because generation requires many full-sequence denoising passes with a large Transformer and, unlike autoregressive decoding, cannot benefit from KV caching. In this work, we exploit the flexibility of the diffusion framework and study model scheduling, where a smaller MDLM replaces the full model at a subset of denoising steps. \rev{Across models trained on OpenWebText and LM1B}, we show that early and late denoising steps are substantially more robust to such replacement than middle steps, enabling up to a 17\% reduction in FLOPs with only modest degradation in generative perplexity \rev{under both unconditional and prefix-conditional generation, while preserving sample diversity}. We support these findings with a step-importance analysis based on loss and KL divergence between small and large models across timesteps, as well as an exhaustive search over coarse step segments, both of which identify the middle of the diffusion trajectory as most sensitive \rev{consistently across datasets}. Our results suggest that simple, architecture-agnostic scheduling rules can significantly accelerate MDLM sampling while largely preserving generation quality.
\end{abstract}

\section{Introduction}

\begin{figure}[t]
    \centering
    \includegraphics[width=1\linewidth]{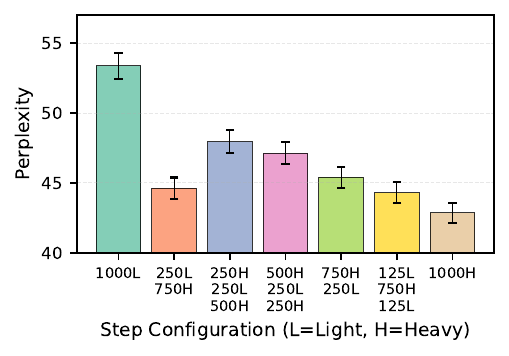}
    \caption{
    Generative perplexity for model schedules using a heavy 12-block model and a light 4-block model with exactly 250/1000 light steps (16.7\% saved FLOPs) on OpenWebText. Each bar label encodes a schedule as contiguous segments, e.g., $(\mathrm{L}125,\mathrm{H}750,\mathrm{L}125)$ denotes the \emph{sandwich} schedule (125 light steps, 750 heavy steps, 125 light steps), while placing all light steps in the 2nd or 3rd quarter yields the worst perplexity. Error bars correspond to 95\% confidence intervals. 
    }
    \label{fig:12b-4b-barplot-perplexity}
\end{figure}

Masked diffusion language models (MDLMs)~\cite{sahoo_simple_2024} have recently emerged as a competitive alternative to autoregressive language models, narrowing the quality gap~\cite{gong_diffucoder_2025,nie_large_2025,ye_dream_2025} while offering a different generation paradigm based on iterative denoising. However, MDLM sampling remains expensive: generation requires many full-sequence denoising passes with a large Transformer, and unlike autoregressive decoding, this process cannot benefit from KV caching~\cite{wu_fast-dllm_2025, wu_fast-dllm_2025-1}. As a result, even when MDLM quality is strong, inference cost can be a practical bottleneck.

A distinctive feature of diffusion models is that generation proceeds through a sequence of timesteps that gradually transform a high-noise (or heavily corrupted) state into a clean sample. This structure suggests a natural question: are all denoising steps equally “difficult,” and therefore equally deserving of full model capacity? In continuous image diffusion, a growing body of work~\cite{shen_md-dit_2024, huang_flexiffusion_2025} explores timestep-dependent compute allocation, including approaches that skip, cache, or dynamically adjust capacity across the trajectory. These methods are motivated by evidence that model behavior varies systematically across timesteps, often exhibiting relatively smooth or monotonic trends in step difficulty. For example, DyDiT++~\cite{zhao_dydit_2025} analyzes loss differences between small and large diffusion Transformers and reports that these differences can shrink toward one end of the trajectory, suggesting that some timesteps may be handled effectively by smaller models.

Whether similar conclusions hold for discrete masked diffusion in language remains unclear. Text denoising differs from image denoising in both the state space (discrete tokens with masking) and the structure of the prediction problem (categorical distributions over vocabularies, with uncertainty concentrated on masked positions). Consequently, step-importance patterns and effective acceleration strategies may not transfer directly from continuous image diffusion to masked diffusion for text.

In this work, we study model scheduling for faster MDLM sampling: at inference time, we replace a subset of denoising steps of a large “heavy” MDLM with a separately trained smaller “light” MDLM. This approach is intentionally simple and architecture-agnostic: it does not require retraining the heavy model, distillation, or modifying the sampling algorithm beyond choosing which model to run at each step. The central question is then straightforward: \textbf{which timesteps are most robust to model replacement, and how should light and heavy steps be arranged to best trade off speed and quality? }

\rev{Our empirical results on OpenWebText and LM1B show that denoising steps are not equally important for masked diffusion generation.} When we fix a compute budget (e.g., replacing 25\% of steps with a light model), the placement of these light steps matters substantially: replacing steps in the middle of the trajectory yields the largest degradation in generative perplexity, while allocating light steps near the beginning and end performs best. In particular, a simple sandwich schedule that places light steps at both ends of the trajectory consistently outperforms schedules that concentrate light steps in the middle. These observations enable meaningful inference savings, achieving up to a 17\% reduction in FLOPs with only modest degradation in generative perplexity.

To validate that this pattern is not an artifact of a small set of hand-designed schedules, we perform an exhaustive search over coarse step segments and find the same qualitative conclusion: middle segments are the most sensitive to replacement, while the earliest and latest segments are relatively safe. This yields a practical rule of thumb: under a fixed budget of cheap steps, it is preferable to distribute them across both ends of the trajectory rather than concentrating them in the middle.

We further support these findings with a step-importance analysis based on \textbf{model similarity vs. timestep}. We compare light and heavy models on the same corrupted inputs at each timestep and measure their disagreement via differences in masked-token cross-entropy and token-level KL divergence. Both measures exhibit a clear peak in the middle of the trajectory, indicating maximal divergence between small and large models at intermediate noise levels. This provides a mechanistic explanation for why middle-step replacement is most harmful and clarifies how step sensitivity in masked diffusion for text differs from the smoother, often monotonic step-importance trends reported in prior continuous image diffusion analyses.

In summary, our contributions are:

\begin{enumerate}
    \item 
Model scheduling for MDLMs: We study an inference-time acceleration strategy that mixes a heavy MDLM with a separately trained light MDLM across denoising steps, without distillation or architecture modification. 
\item
\rev{Empirical step-importance finding: Across two datasets (OpenWebText and LM1B) and both unconditional and prefix-conditional generation, we show that early and late denoising steps are substantially more robust to model replacement than middle steps. This yields a continuous quality--efficiency tradeoff controlled by light-model size and step fraction (e.g., from 3.4\% perplexity degradation at 16.7\% FLOPs savings to larger reductions at higher substitution rates), while preserving sample diversity.}

\item
\rev{Explanatory analysis: We provide complementary evidence from (i) loss/KL-based similarity across timesteps and (ii) exhaustive search over coarse segments, both identifying the middle of the trajectory as most compute-sensitive. This peaked pattern contrasts with the monotonic trends reported in continuous image diffusion, revealing a qualitatively different step-importance structure in discrete masked diffusion for text.}
\end{enumerate}

Section~\ref{sec:related_work} reviews masked diffusion LMs and related efficiency work; Section~\ref{sec:experiments} introduces model scheduling; Section~\ref{sec:analsys} presents empirical results and step-importance analyses; Section~\ref{sec:conclusion} discusses limitations and future directions.

\section{Related Work}
\label{sec:related_work}

\subsection{Diffusion Models}
Denoising diffusion probabilistic models (DDPMs) \cite{ho_denoising_2020} and score-based generative models \cite{song_score-based_2021} have become a standard framework for high-fidelity generation. Beyond the original ancestral samplers, a large body of work has improved sampling efficiency via alternative discretizations and solvers \cite{song_denoising_2022,lu_dpm-solver_2022}. For high-capacity backbones, diffusion transformers (DiT) \cite{peebles_scalable_2023} and related Transformer-based score/denoiser parameterizations dominate modern image diffusion systems~\cite{rombach_high-resolution_2022}.

\subsection{Combining Diffusion Models}
A classical way to combine models at sampling time is guidance, including classifier guidance \cite{dhariwal_diffusion_2021} and classifier-free guidance (CFG) \cite{ho_classifier-free_2022}. Closer to our setting, several vision works explicitly mix models of different sizes across the denoising trajectory to trade off speed and quality without retraining: OMS-DPM \cite{liu_oms-dpm_2023} searches for an optimal per-timestep model assignment under a time budget, while T-Stitch \cite{pan_stitched_2023} “stitches” a small model into the early part of the trajectory as a drop-in replacement.

\subsection{Diffusion Models Acceleration}
Diffusion acceleration methods fall into two broad categories: reducing the number of function evaluations (e.g., DDIM \cite{song_score-based_2021}, DPM-Solver \cite{lu_dpm-solver_2022}) and reducing the cost of each evaluation (distillation/consistency and architecture-level adaptivity). Distillation-based methods such as progressive distillation \cite{salimans_progressive_2022} and consistency models \cite{song_consistency_2023} aim to preserve quality with fewer steps.
A complementary direction makes the \emph{denoiser itself} compute-adaptive. Step-aware or schedule-aware diffusion backbones include DDSM \cite{yang_denoising_2024}, which studies step importance and step-dependent capacity, as well as diffusion-transformer specific techniques that skip/caches computation or dynamically route capacity, such as Learning-to-Cache \cite{ma_learning--cache_2024}, Dynamic Diffusion Transformers (DyDiT) \cite{zhao_dydit_2025}, MD-DiT \cite{shen_md-dit_2024}, AdaDiff~\cite{tang_adadiff_2024}. Related NAS-style methods (e.g., Flexiffusion \cite{huang_flexiffusion_2025}) also optimize which segments are full/cached/skipped to meet a compute target. These works are primarily developed and evaluated in continuous image diffusion, and their conclusions about which timesteps deserve more capacity do not necessarily transfer to discrete masked diffusion for text.

\subsection{Masked Diffusion Language Models}
Diffusion for text has been explored in both continuous and discrete spaces. Continuous-text diffusion includes Diffusion-LM \cite{DBLP:conf/nips/LiTGLH22}, latent variants~\cite{lovelace_latent_2023}, and conditional variants such as DiffuSeq \cite{gong_diffuseq_2023} and simplex-based~\cite{meshchaninov_compressed_2025, shabalin_smoothie_2025}. Discrete diffusion models for language build on discrete-state diffusion formulations such as D3PM \cite{austin_structured_2023}, and include DiffusionBERT \cite{he_diffusionbert_2022} and Score Entropy Discrete Diffusion (SEDD) \cite{lou_discrete_2024}. Recent masked diffusion language models (MDLMs) \cite{sahoo_simple_2024, shi_simplified_2025} show that a simple masked diffusion objective with strong training recipes can close much of the quality gap to autoregressive LMs, and ReMDM \cite{wang_remasking_2025} improves sampling via inference-time remasking and compute scaling. Compared to AR LMs, diffusion LMs can also be stronger learners under data-constrained settings~\cite{ni_diffusion_2025, rutte_scaling_2025}.

Several efforts scale diffusion LMs and explore hybridization with autoregressive decoding, including large-scale MDLM reports such as LLaDA \cite{nie_large_2025} and Dream \cite{ye_dream_2025}, as well as domain/architecture variants such as DiffuCoder~\cite{gong_diffucoder_2025} and DiffuLLaMA~\cite{nie_scaling_2025}. A separate line of work targets inference efficiency. Some methods recover or approximate KV-cache benefits for bidirectional diffusion via block/hybrid formulations and cache reuse \cite{arriola_block_2025, wu_fast-dllm_2025, sahoo_esoteric_2025, arriola_encoder-decoder_2025, wu_fast-dllm_2025-1}. Others reduce the effective number of denoising iterations through adaptive or distilled decoding policies, including FlashDLM (FreeCache and AR-guided step reduction) \cite{hu2025flashdlm0}, LocalLeap~\cite{kong2025accelerating}, and CD4LM~\cite{liang2026cd4lm0}. Finally, dInfer provides a system-oriented framework with modular decoding strategies and KV-cache management for efficient diffusion-LM serving \cite{ma2025dinfer0}. These approaches are largely \emph{orthogonal} to our model scheduling: they reduce the number of denoising iterations, the per-step attention cost via caching, and/or system overhead, whereas we vary \emph{model capacity across steps} without modifying the sampler. In principle, scheduling composes with KV caching (apply caching within both heavy and light steps) and with step-reduction decoders (apply capacity scheduling within the remaining iterations), suggesting multiplicative speedups.

Finally, token difficulty is known to be non-uniform in autoregressive generation, with evidence that per-position perplexity can vary systematically across a sequence \cite{helm_token_2025, zur_are_2025, yang_llm_2025, bell_slaves_2025}; this motivates exploring whether “where to spend compute” is similarly non-uniform across diffusion timesteps in masked diffusion generation.

\section{Accelerating MDLM via Model Scheduling}
\label{sec:experiments}

\subsection{Experimental setup}
\label{sec:setup}

\paragraph{Masked diffusion language models.}
Let $x = (x^{(1)},\dots,x^{(L)})$ denote a clean (denoised) token sequence of length $L$. Autoregressive language models generate $x$ sequentially by modeling $p_\theta(x^{(i)} \mid x^{(<i)})$ and are typically trained with a token-level cross-entropy objective. In contrast, masked diffusion language models (MDLMs) generate text by repeatedly denoising a partially masked sequence. Concretely, we define a discrete forward noising process $q$ that corrupts $x$ into $z_t$ by replacing tokens with a special mask token $m$ according to a time-dependent corruption level.

\paragraph{Forward process $q(z_t \mid x)$.}
We represent each token as a one-hot vector in $\{0,1\}^{|V|}$. For a normalized time $t \in [0,1]$, the forward process produces a noisy sequence $z_t = (z_t^{(1)},\dots,z_t^{(L)})$ with independent per-position marginals
\begin{equation}
q(z_t^{(\ell)} \mid x^{(\ell)}) = \mathrm{Cat}\!\left( \alpha_t \, x^{(\ell)} + (1-\alpha_t)\,\pi \right),
\label{eq:forward_cat}
\end{equation}
where $\pi$ is a fixed prior distribution over vocabulary tokens. In our setting we use \emph{pure masking}, i.e., $\pi=\delta_m$, so that with probability $(1-\alpha_t)$ a token is replaced by $m$, and otherwise it is kept unchanged. We use a \emph{linear} schedule $\alpha_t = 1-t$, so that the expected masked fraction equals $t$. During training we sample $t \sim \mathcal{U}(0,1)$.

Let $M(z_t) \subseteq \{1,\dots,L\}$ denote the set of masked positions in $z_t$, i.e., $M(z_t)=\{\ell : z_t^{(\ell)} = m\}$.

\paragraph{Denoiser and training objective.}
The denoiser $p_\theta(x \mid z_t, t)$ is parameterized by a bidirectional Transformer that predicts the original token at each position given the noisy sequence and timestep. Training minimizes a weighted masked language modeling loss over masked positions only. Following prior derivations of the (negative) ELBO for this discrete diffusion process, the objective can be written as
\begin{equation}
\label{eq:mdlm_loss}
\begin{aligned}
\mathcal{L}_{\mathrm{MDLM}}(x)
&=
\mathbb{E}_{t \sim \mathcal{U}(0,1),\, z_t \sim q(\cdot \mid x)}
\left[ \ell_\theta(x; z_t, t) \right], \\
\ell_\theta(x; z_t, t)
&=
\frac{\alpha'_t}{1-\alpha_t}
\sum_{\ell \in M(z_t)}
-\log p_\theta\!\left(x^{(\ell)} \mid z_t, t\right).
\end{aligned}
\end{equation}
Here $\alpha'_t$ denotes the derivative of $\alpha_t$ with respect to $t$. The factor $\frac{\alpha'_t}{1-\alpha_t}$ reweights timesteps so that different corruption levels contribute appropriately to the variational objective; for $\alpha_t = 1-t$, we have $\frac{-\alpha'_t}{1-\alpha_t} = \frac{1}{t}$.

\paragraph{Sampling (reverse process).}
To generate an unconditional sequence of length $L$, sampling starts from a fully masked sequence $z_{t=1}$ where $z^{(\ell)}_{t=1}=m$ for all $\ell$. The reverse process proceeds for $T$ discrete steps with times $t_i=i/T$ for $i=T,T-1,\dots,0$. We use the standard MDLM sampler (no remasking): once a token is generated (unmasked), it remains fixed thereafter. Multiple tokens can be updated in parallel at each step (unlike autoregressive decoding); however, each denoising step requires a full bidirectional Transformer forward pass over the entire sequence, and thus inference cost scales with the number of denoising evaluations.

\paragraph{Why sampling is expensive.}
Although MDLM sampling updates many tokens at once, it typically requires a large number of sequential denoising steps and does not admit the KV-caching efficiency of autoregressive decoding. This motivates our focus on \emph{model scheduling}: replacing the full denoiser with a smaller denoiser on a subset of timesteps to reduce total compute while retaining generation quality.

\paragraph{Model scheduling.}
Let $\{\theta_k\}_{k \in \mathcal{K}}$ denote a set of denoisers of different sizes (e.g., $k\in\{4,6,8,10,12\}$ Transformer blocks) trained with the same objective and noise schedule.
A \emph{model schedule} is a function $s:\{1,\dots,T\}\to\mathcal{K}$ that selects which denoiser to use at each reverse step $i$ (time $t_i=i/T$). Sampling then applies $p_{\theta_{s(i)}}(\cdot \mid z_{t_i}, t_i)$ at step $i$.
If the heavy model has $B_H=12$ blocks and the light model has $B_L$ blocks, then replacing a fraction $p$ of steps by the light model yields a relative compute reduction of
\begin{equation}
\text{saved FLOPs} \approx p \cdot \frac{B_H - B_L}{B_H}.
\label{eq:saved_flops}
\end{equation}

\paragraph{Models and training details.}
To avoid confounding factors and isolate the effect of scheduling, we closely follow the MDLM training setup~\cite{sahoo_simple_2024} and use their codebase and default design choices whenever possible. We train a family of Transformer-encoder denoisers~\cite{vaswani2017attention,DBLP:conf/naacl/DevlinCLT19} that differ \emph{only} in depth (4/6/8/10/12 blocks) while keeping width fixed (hidden size 768, MLP ratio 4, same vocabulary/tokenizer). The 12-block model serves as the \emph{heavy} baseline, and the smaller models serve as candidate \emph{light} denoisers. Because Transformer blocks are executed sequentially, both runtime and FLOPs scale approximately linearly with the number of blocks, enabling simple and reliable compute accounting for our schedules.

All models are trained on OpenWebText~\cite{Gokaslan2019OpenWeb} tokenized with the GPT-2 tokenizer~\cite{brown2020language} for 1M optimization steps with effective batch size 512 and sequence length 1024. This corresponds to approximately 262B masked tokens during training. We choose OpenWebText as a broad, general-purpose natural language corpus to study unconditional generation and isolate the effect of timestep scheduling without task-specific structure. We use AdamW~\cite{loshchilov2017decoupled} with 2500 linear warmup steps, learning rate $3\cdot10^{-4}$, and $\beta=(0.9,0.999)$ (other hyperparameters follow~\cite{sahoo_simple_2024}).

\paragraph{Evaluation metric.}
To measure unconditional generation quality, we follow MDLM~\cite{sahoo_simple_2024} and report \emph{generative perplexity} computed by a pretrained GPT-2~\cite{brown2020language} model on fully unconditional samples. Unless stated otherwise, we generate 1600 independent samples of length 1024 using $T=1000$ denoising steps and compute mean perplexity. We acknowledge that generative perplexity can be unreliable in some settings (e.g., ReMDM~\cite{wang_remasking_2025} discusses failure modes), but in this work we compare schedules under identical training and sampling protocols, and use it as a consistent \emph{relative} metric across configurations. \rev{We additionally report token-level entropy as a sample diversity measure and evaluate under prefix-conditional generation (Section~\ref{sec:prefix_cond}).}

\rev{
\paragraph{Second dataset.} To test whether our findings generalize beyond a single corpus, we train an identical model family on the One Billion Word Benchmark (LM1B)~\cite{chelba2014lm1b} with 128-token sequence length. All other architecture and training choices remain the same.
}

\subsection{Fixed light-step ratio (25\%)}
\label{sec:fixed25}

We first consider a simple setting with two models: a 12-block heavy MDLM and a 4-block light MDLM. We replace exactly 25\% of the heavy model's denoising steps with the light model and ask: \emph{which steps should be replaced to minimize quality loss?} Under our compute accounting in Eq.~\ref{eq:saved_flops}, the saved FLOPs are $16.7\%$.
Although our framework naturally extends to using more than two models, we focus on this clear two-model setup to make the effect of schedule placement easy to interpret.

We test several hand-crafted schedules that place the 250 light steps in different parts of the trajectory (by quarters), and also a \emph{sandwich} schedule that splits the 250 light steps into two equal segments of 125 and places them at the beginning and end of the trajectory. The results are shown in Figure~\ref{fig:12b-4b-barplot-perplexity}. Replacing steps in the middle of the trajectory (2nd/3rd quarters) yields the worst perplexity, while the sandwich schedule performs best, closely followed by placing all light steps in the first quarter. These results indicate that denoising steps are not equally important for masked diffusion generation. \rev{Token-level entropy for these schedules (Table~\ref{tab:schedule_ppl_entopy_owt_12b-4b} in Appendix~\ref{app:entropy}) confirms stable sample diversity across all configurations.} Additional hand-crafted schedules for other light model sizes are reported in Appendix~\ref{app:more_configs} and exhibit the same qualitative pattern.
\rev{The same pattern holds on our LM1B models (Appendix~\ref{app:lm1b}, Figure~\ref{fig:lm1b-12b-4b-barplot-perplexity}), confirming cross-dataset generality.}

\subsection{Exhaustive search over coarse step segments}
\label{sec:bruteforce}

To further validate the above trend under a stronger compute reduction, we replace 400 out of 1000 denoising steps (40\%) with the 4-block light model. This corresponds to $26.7\%$ saved FLOPs. While some prior works search over timesteps via learned predictors~\cite{liu_oms-dpm_2023} or heuristic optimization schemes~\cite{yang_denoising_2024}, we perform an exhaustive search in a discretized space for transparency.

A naive search over all subsets of 400 steps is infeasible: $\binom{1000}{400}\approx 5\times 10^{290}$. We therefore partition the 1000 steps into 10 contiguous segments of 100 steps and select 4 segments to run with the light model, resulting in $\binom{10}{4}=210$ schedules. For this brute-force experiment, we evaluate each schedule using 160 unconditional samples (fixed seeds across schedules) for tractability.

\begin{figure}
    \centering
    \includegraphics[width=0.95\linewidth]{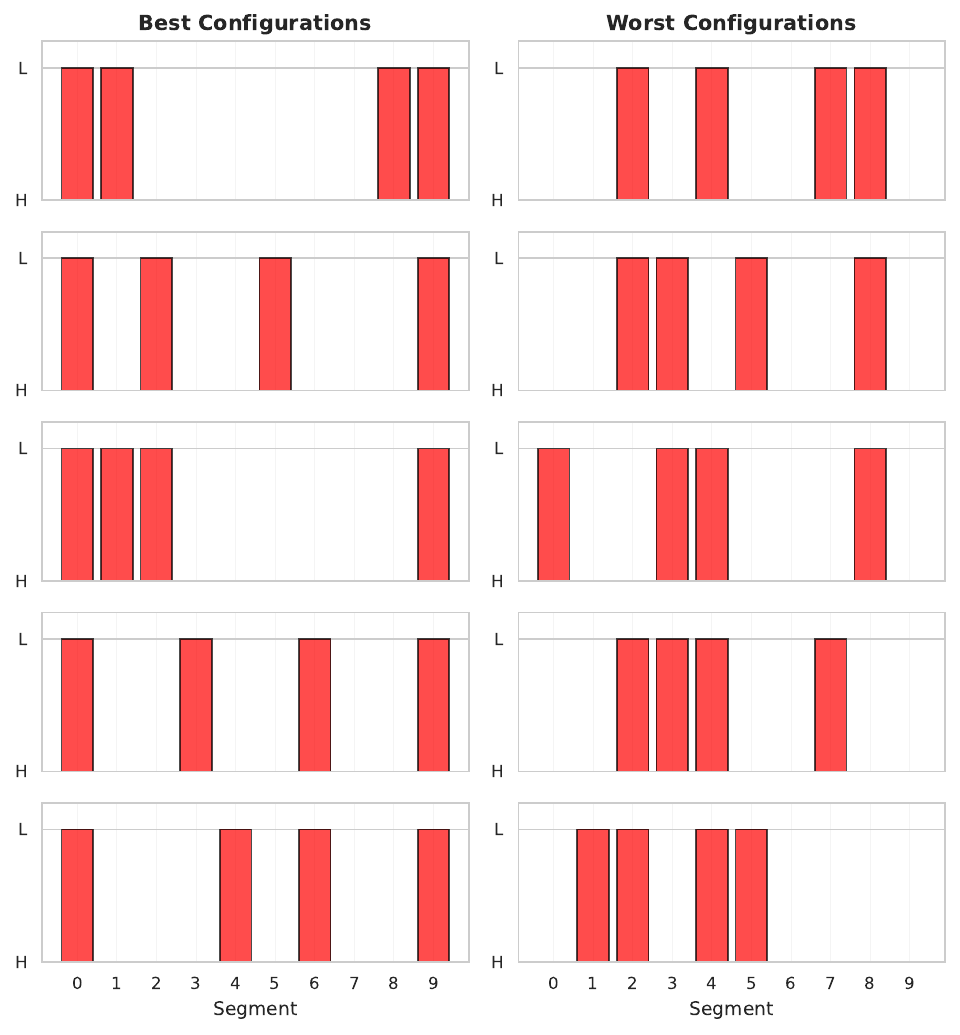}
    \caption{Comparison of the top 5 best (left) and worst (right) model scheduling configurations among the 210 coarse schedules. Each row shows one configuration. Red bars indicate light (4-block) model placement. Segments 0--9 correspond to steps 0--100, 100--200, \dots, 900--1000, where segment 0 is closest to the fully masked state ($t\approx 1$). Best configurations concentrate light segments near both ends, while worst configurations place light segments in the middle.}
    \label{fig:best_worst_barplot}
\end{figure}

Figure~\ref{fig:best_worst_barplot} compares the top-5 and bottom-5 schedules. The best schedules consistently place light segments near the beginning and end of the trajectory, while the worst schedules place light segments predominantly in the middle.
We quantify this by counting segment frequency among the top-20 and bottom-20 schedules (Figure~\ref{fig:top20config_barplot}; bottom-20 in Appendix~\ref{app:bottom20}, Figure~\ref{fig:bottom20config_barplot}). Middle segments appear disproportionately often in the worst schedules, confirming that mid-trajectory steps are the most sensitive to replacement.

\paragraph{Implementation note.}
This coarse segmentation is also convenient in practice: the MDLM sampler can include ``no-op'' iterations where the mask set does not change, allowing us to reuse logits instead of re-running the Transformer. Using contiguous segments simplifies this bookkeeping (this is not autoregressive KV caching).

\begin{figure}
    \centering
    \includegraphics[width=0.95\linewidth]{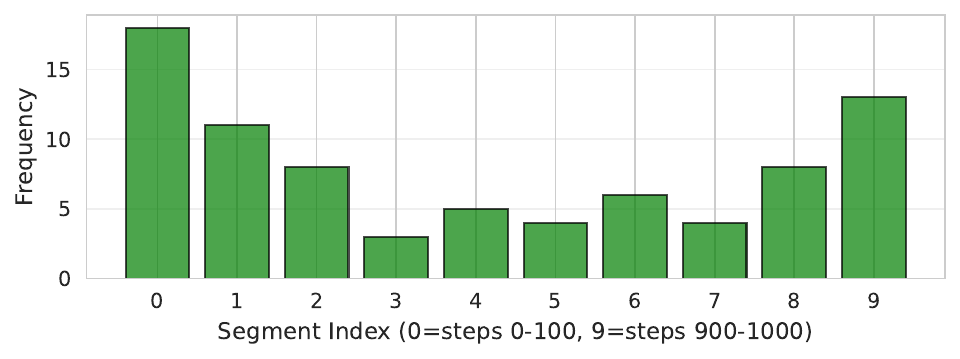}
    \caption{Segment frequency in the top 20 best-performing configurations (lowest perplexity). Bars show how often each segment is assigned to the light (4-block) model across the top-20 schedules. Higher frequency suggests that replacing this segment is relatively safe.}
    \label{fig:top20config_barplot}
\end{figure}

As a practical rule of thumb suggested by both the hand-crafted and exhaustive results, spreading cheaper steps across \emph{both} ends of the trajectory tends to be preferable to concentrating them in the middle. For example, for 600 light steps one can use a symmetric schedule of $(\mathrm{L}300,\mathrm{H}400,\mathrm{L}300)$.

\subsection{Scaling over light model size / light-step fraction}
\label{sec:scaling}

We next study two scaling dimensions: (i) the size of the light model and (ii) the fraction of light steps.
Table~\ref{tab:scaling_models} fixes the \emph{sandwich} placement (125 light + 750 heavy + 125 light) and varies the light model depth from 4 to 10 blocks, paired with the 12-block heavy baseline. As expected, increasing light-model depth reduces the quality drop while also reducing the achievable FLOPs savings.

Table~\ref{tab:scaling_steps} fixes the model pair (12-block heavy, 4-block light) and varies the percentage of steps executed by the light model from 0\% to 100\%. In addition to FLOPs-based estimates, we report end-to-end wall-clock time for the same sampling setup. We observe a smooth transition in perplexity as the schedule shifts from fully heavy to fully light, indicating that model scheduling provides a continuous speed--quality tradeoff.

\begin{table}[h]
    \centering
    \caption{Scaling the \emph{light model} number of \textbf{b}locks while keeping schedule placement fixed to the sandwich pattern (125 light + 750 heavy + 125 light, i.e., 25\% light steps). ``PPL drop'' is relative to the all-heavy 12-block baseline. ``Saved FLOPs'' is computed as $0.25\cdot(12-B_L)/12$. Numbers in {\scriptsize{scriptsize}} represent 95\% confidence intervals.}
    \label{tab:scaling_models}

\begin{tabular}{lccc}
\toprule
Light Model & Gen.\ PPL & PPL drop & Saved FLOPs \\
\midrule
4b  & 44.31{\scriptsize\,$\pm$\,0.76} & 3.41\% & 16.67\% \\
6b  & 43.67{\scriptsize\,$\pm$\,0.67} & 1.94\% & 12.50\% \\
8b  & 43.45{\scriptsize\,$\pm$\,0.73} & 1.40\% & 8.33\%  \\
10b & 42.90{\scriptsize\,$\pm$\,0.70} & 0.12\% & 4.17\%  \\
12b & 42.85{\scriptsize\,$\pm$\,0.71} & 0.00\% & 0.00\%  \\
\bottomrule
\end{tabular}

\end{table}

\begin{table}[t]
    \centering
    \caption{Scaling the \emph{fraction of light steps} for the (12-block heavy, 4-block light) model pair under a `sandwhich` pattern.}
    \label{tab:scaling_steps}
    % \small
    \setlength{\tabcolsep}{4pt}
    \begin{tabular}{r r r r r}
        \toprule
        \makecell{\% light\\steps} & Gen.\ PPL & Saved FLOPs & Time (s) & Speedup \\
        \midrule
        0   & 42.9 & 0.0\%  & 109.7 & 0.0\% \\
        \midrule
        10  & 43.1 & 6.7\%  & 106.4 & 3.0\% \\
        20  & 43.8 & 13.3\% & 103.3 & 5.8\% \\
        30  & 44.7 & 20.0\% & 100.4 & 8.5\% \\
        40  & 45.9 & 26.7\% & 97.3  & 11.3\% \\
        50  & 47.2 & 33.3\% & 94.3  & 14.0\% \\
        60  & 48.6 & 40.0\% & 91.1  & 17.0\% \\
        70  & 50.1 & 46.7\% & 90.7  & 17.3\% \\
        80  & 51.4 & 53.3\% & 85.4  & 22.2\% \\
        90  & 52.5 & 60.0\% & 84.8  & 22.7\% \\
        \midrule
        100 & 53.4 & 66.7\% & 78.7  & 28.3\% \\
        \bottomrule
    \end{tabular}
\end{table}

\paragraph{Wall-clock vs.\ FLOPs.}
Although our compute estimates scale linearly with Transformer depth, measured wall-clock speedups (Table~\ref{tab:scaling_steps}) may differ because not all inference cost is depth-dependent. In our MDLM implementation, the input/output embedding and, in particular, the final vocabulary projection dominate runtime for smaller models, and these layers are identical across the heavy and light variants. Profiling shows this: for the 4-block model, the output layer accounts for the majority of runtime ($\approx 81.6\%$), whereas the Transformer blocks contribute only $\approx 18.2\%$; for the 12-block model, the output layer remains substantial ($\approx 59.9\%$) while blocks account for $\approx 40.0\%$. As a result, reducing depth primarily affects the block compute and yields smaller end-to-end speedups than predicted by FLOPs alone. This mismatch should shrink at larger scales or when block compute dominates (e.g., larger hidden sizes, longer sequences, or architectures where the output projection is relatively less significant). Thus, FLOPs savings should be viewed as an upper bound on wall-clock gains unless the non-depth-dependent components (e.g., vocabulary projection and softmax/sampling) are also optimized. Importantly, this bottleneck is not fundamental: more efficient implementations for the output projection and softmax/sampling exist (e.g., fused projection--softmax/loss operators as in Liger-Kernel~\cite{hsu2025ligerkernel}, and highly optimized inference kernels in serving stacks such as NVIDIA TensorRT-LLM~\cite{nvidia_tensorrtllm_docs} and FlashInfer~\cite{ye2025flashinfer0}). Leveraging such kernels is orthogonal to our scheduling method and can both increase the absolute speedups and bring wall-clock gains closer to FLOPs-based predictions.

\noindent\textbf{A simple model.}
If the end-to-end runtime decomposes as $T \approx T_{\text{out}} + T_{\text{blocks}}$, and only $T_{\text{blocks}}$ scales with depth, then the attainable speedup from reducing depth is limited by the fraction of time spent outside the blocks. Writing $\alpha \coloneqq T_{\text{out}}/T$, the maximum possible speedup satisfies \( \mathrm{speedup} \;\le\; \frac{1}{1-\alpha} \).

\subsection{\rev{Prefix-conditional generation and diversity}}
\label{sec:prefix_cond}
\rev{
We also test whether the same scheduling pattern holds under prefix-conditional generation, which is more relevant in practice. We repeat the schedule comparison using prefix-conditional sampling: generation starts from 256 prefix tokens drawn from held-out OpenWebText data, and the remaining positions are denoised.
Table~\ref{tab:owt_schedule_ppl_prefix256} reports conditional generative perplexity and token-level entropy. The schedule ranking is unchanged: middle-step replacement gives the worst perplexity, while the sandwich schedule remains close to the all-heavy baseline (40.9 vs.\ 39.3). Entropy stays stable across all schedules (5.40--5.42), suggesting that scheduling does not reduce sample diversity. The same stability appears in unconditional generation (Appendix~\ref{app:entropy}, Table~\ref{tab:schedule_ppl_entopy_owt_12b-4b}). Results with 128-token prefixes show the same pattern (Appendix~\ref{app:prefix_cond}, Figure~\ref{fig:owt_perplexity_barplot_12b4b_prefix128}).
}
\begin{table}[t]
\centering
\begin{tabular}{@{}l r r@{}}
\toprule
Schedule & Cond.\ Gen.\ PPL {\scriptsize (CI)} & Entropy \\
\midrule
L1000 & 47.5 {\scriptsize\,$\pm$\,0.44} & 5.42 \\
L250 $\to$ H750 & 40.9 {\scriptsize\,$\pm$\,0.38} & 5.42 \\
H250 $\to$ L250 $\to$ H500 & 43.2 {\scriptsize\,$\pm$\,0.41} & 5.41 \\
H500 $\to$ L250 $\to$ H250 & 42.6 {\scriptsize\,$\pm$\,0.41} & 5.41 \\
H750 $\to$ L250 & 41.4 {\scriptsize\,$\pm$\,0.40} & 5.40 \\
L125 $\to$ H750 $\to$ L125 & 40.9 {\scriptsize\,$\pm$\,0.36} & 5.42 \\
H1000 & 39.3 {\scriptsize\,$\pm$\,0.37} & 5.40 \\
\bottomrule
\end{tabular}
\caption{Prefix-conditional generation (256-token prefix from OpenWebText). Schedule ranking matches the unconditional setting (Figure~\ref{fig:12b-4b-barplot-perplexity}). \textbf{H}eavy: 12-block, \textbf{L}ight: 4-block. Entropy remains stable across schedules, indicating no diversity loss from scheduling.}
\label{tab:owt_schedule_ppl_prefix256}
\end{table}

\section{Why does this work? Step importance analysis}
\label{sec:analsys}

\subsection{Model similarity vs timestep}
\label{sec:similarity}

Inspired by OMS-DPM~\cite{liu_oms-dpm_2023}, we compare model behavior across noise levels. Following the similarity analyses used in dynamic diffusion transformer works (e.g.,~\cite{zhao_dydit_2025}), we compute loss differences and KL divergences between models at fixed timesteps. For each timestep we evaluate on 500 sequences of length 1024, and crucially compare models on the \emph{same} corrupted inputs.

\begin{figure}
    \centering
    \includegraphics[width=0.95\linewidth]{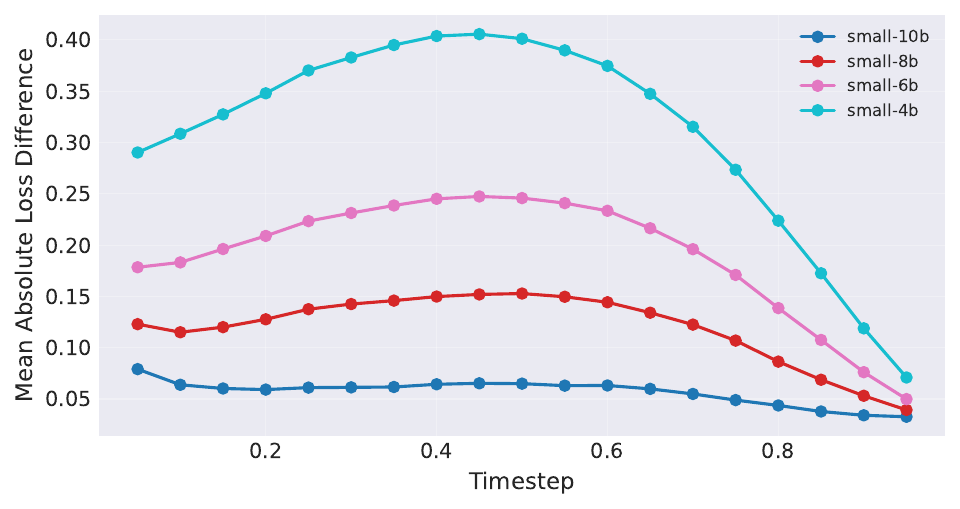}
    \caption{Mean absolute difference in masked-token cross-entropy between each light model and the heavy 12-block baseline across timesteps. Each curve compares one light model to the baseline, evaluated on the same corrupted inputs $z_t$. Lower values indicate higher similarity.}
    \label{fig:loss_diff}
\end{figure}

\paragraph{Loss difference.}
For a fixed timestep $t$, we sample $z_t\sim q(\cdot\mid x)$ and compute the (unweighted) masked-token cross-entropy
\begin{equation}
\mathcal{L}_\theta(z_t,t) = \frac{1}{|M(z_t)|}\sum_{\ell\in M(z_t)} -\log p_\theta(x^{(\ell)}\mid z_t,t).
\end{equation}
We then measure the mean absolute loss difference between a light model $\theta_L$ and the heavy model $\theta_H$:
\begin{equation}
\Delta_{\mathrm{loss}}(t) =
\mathbb{E}_{x}\,\mathbb{E}_{z_t\sim q(\cdot\mid x)}
\left|
\mathcal{L}_{\theta_L}(z_t,t) - \mathcal{L}_{\theta_H}(z_t,t)
\right|.
\label{eq:loss_diff}
\end{equation}
The result is presented in Figure~\ref{fig:loss_diff}. Prior work on continuous image diffusion often reports a monotonic trend across timesteps~\cite{pan_t-stitch_2024,zhao_dydit_2025}. In contrast, we observe a clear peak in the middle of the trajectory, indicating maximal disagreement between light and heavy models at intermediate noise levels.

\begin{figure}
    \centering
    \includegraphics[width=1\linewidth]{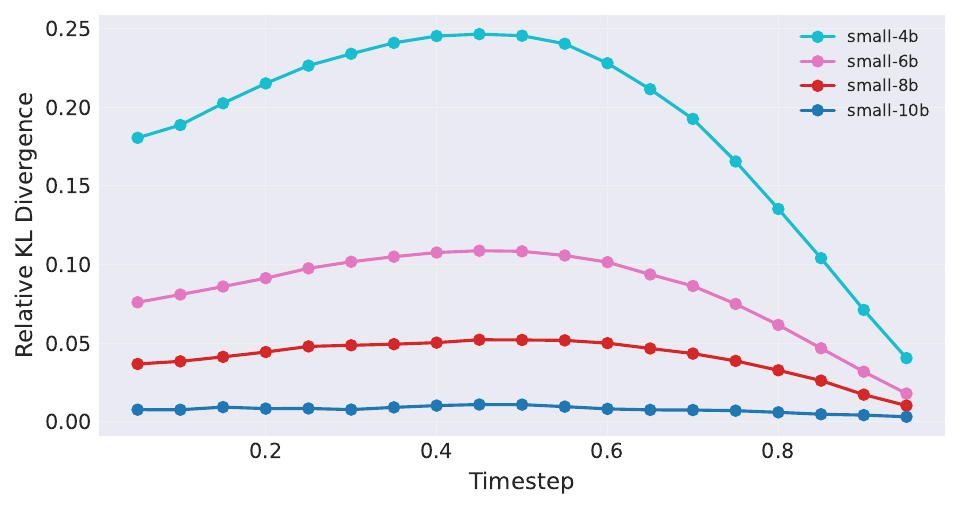}
    \caption{Relative token-level KL divergence (Eq.~\ref{eq:token_kl}) between model pairs across timesteps, after subtracting a baseline KL curve computed between two independently trained heavy (12-block) checkpoints. Lower values indicate closer agreement. Divergence peaks in the middle of the trajectory, suggesting that intermediate timesteps are most sensitive to model replacement.}
    \label{fig:kldiv-plot}
\end{figure}

\paragraph{KL divergence.}
We additionally compare the \emph{token distributions} predicted at masked positions. Let $p_\theta(\cdot \mid z_t,t,\ell)$ denote the categorical distribution over the vocabulary for position $\ell\in M(z_t)$. 
We compute the average token-level KL divergence between two models:
\begin{equation}
\begin{aligned}
\Delta_{\mathrm{KL}}(t)
&= \mathbb{E}_{x}\,\mathbb{E}_{z_t\sim q(\cdot\mid x)}
\Bigg[\frac{1}{|M(z_t)|}\times \\
&\sum_{\ell\in M(z_t)}
\mathrm{KL} \bigl(
p_{\theta_H}(\cdot \mid z_t,t,\ell)\,\|\,
 p_{\theta_L}(\cdot \mid z_t,t,\ell)
\bigr)\Bigg].
\end{aligned}
\label{eq:token_kl}
\end{equation}
To account for intrinsic ambiguity in text prediction, we compute a baseline $\Delta_{\mathrm{KL}}^{\mathrm{base}}(t)$ as the KL divergence between two independently trained heavy (12-block) checkpoints with different random seeds (different initialization and data order), and report the \emph{relative} divergence $\Delta_{\mathrm{KL}}(t)-\Delta_{\mathrm{KL}}^{\mathrm{base}}(t)$.
Figure~\ref{fig:kldiv-plot} shows the same qualitative pattern as Figure~\ref{fig:loss_diff}: disagreement peaks near the middle ($t\approx 0.4$--$0.6$) and is substantially smaller at both ends. Here $t=1$ corresponds to the fully masked state (start of sampling), and $t\to 0$ corresponds to the nearly unmasked state (end of sampling).

\rev{We replicate this KL-divergence analysis on our LM1B models and observe the same characteristic middle-trajectory peak (Appendix~\ref{app:lm1b}, Figure~\ref{fig:lm1b-kldiv-plot}), confirming that the non-monotonic step-importance pattern is a general property of masked diffusion rather than an artifact of the OpenWebText setup.}

\subsection{Segment influence from exhaustive search}
\label{sec:segment_influence}

\begin{figure}
    \centering
    \includegraphics[width=1\linewidth]{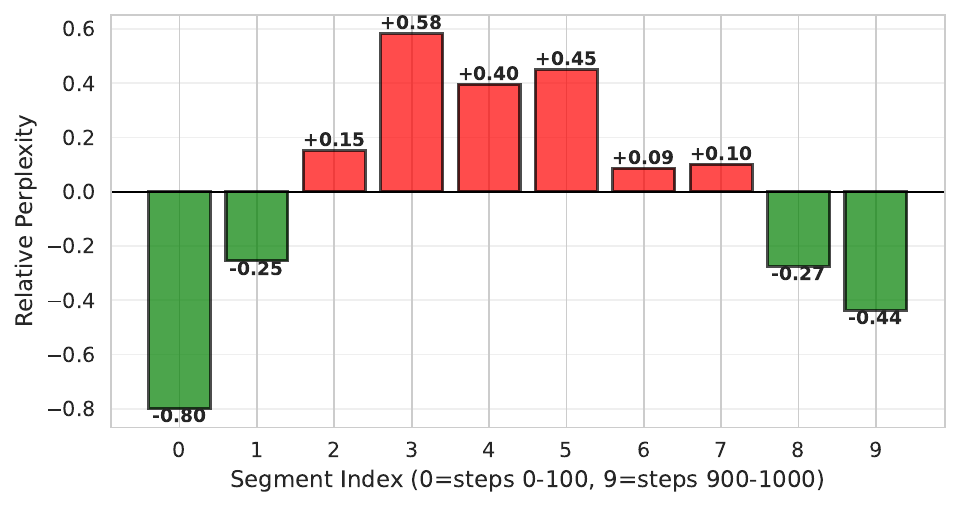}
    \caption{Mean-subtracted segment influence from the exhaustive 10-segment search (Section~\ref{sec:bruteforce}). For each segment $j$, we compute the mean perplexity over schedules that assign segment $j$ to the light model, and subtract the mean perplexity over all schedules. Positive values indicate that replacing this segment is harmful on average; negative values indicate that replacing this segment is relatively safe.}
    \label{fig:per_segment_ppl_barplot}
\end{figure}

The exhaustive 10-segment search in Section~\ref{sec:bruteforce} provides another way to estimate step importance.
For each segment $j\in\{0,\dots,9\}$, let $\mathcal{S}_j$ be the set of schedules in which segment $j$ uses the light model, and let $\mathrm{PPL}(s)$ be the generative perplexity of schedule $s$.
We compute the segment score as
\begin{equation}
I(j) = \frac{1}{|\mathcal{S}_j|}\sum_{s\in\mathcal{S}_j}\mathrm{PPL}(s) \;-\; \frac{1}{|\mathcal{S}|}\sum_{s\in\mathcal{S}}\mathrm{PPL}(s),
\label{eq:segment_score}
\end{equation}
i.e., the mean perplexity of schedules that use the light model in segment $j$, mean-subtracted by the average perplexity over all 210 schedules.
Figure~\ref{fig:per_segment_ppl_barplot} shows that middle segments have positive scores (worse than average when replaced), while the earliest and latest segments tend to have negative scores (more robust to replacement). This empirically matches the similarity analysis above and supports the conclusion that intermediate denoising steps are the most compute-sensitive for masked diffusion generation.

\section{Conclusion}
\label{sec:conclusion}
Masked diffusion language models offer a compelling alternative to autoregressive generation, but their practicality is often constrained by expensive sampling that requires many full-sequence denoising passes. In this work we studied model scheduling for masked diffusion LMs: replacing a subset of denoising steps of a heavy model with a separately trained lighter Transformer at inference time. Our results show that timestep importance in masked diffusion for text is strongly non-uniform: intermediate timesteps are the most sensitive to model replacement, while early and late steps are comparatively robust. This enables simple schedules, such as sandwich-style allocation of light steps to the ends of the trajectory, to reduce sampling compute with only modest degradation in generative perplexity\rev{, while preserving sample diversity as measured by token-level entropy}. \rev{The finding is consistent across two datasets (OpenWebText and LM1B), unconditional and prefix-conditional generation, and multiple analysis methods (schedule search, loss difference, KL divergence).} Notably, this \rev{peaked} pattern contrasts with \rev{the monotonic trends reported in} continuous image diffusion\rev{, indicating a qualitatively different step-importance structure in discrete masked diffusion for text}.

\rev{We emphasize that the main contribution of this work is the empirical identification of non-uniform step importance in masked diffusion for text. Model scheduling serves both as a simple inference-time acceleration baseline and as an experimental tool for exposing this structure. Because it changes per-step model capacity without modifying the sampler, it is complementary to methods that reduce the number of denoising iterations or recover KV-cache-like efficiency, and can in principle compose with them.}

Several natural extensions follow. Currently, pre-trained families of MDLMs spanning multiple scales are not yet standard in the way they are for autoregressive LMs (e.g., Qwen~\cite{yang2025qwen3} or LLaMA~\cite{touvron2023llama}). When such families become available, it will be important to verify our findings at larger scale using established benchmarks. Second, scheduling can be generalized beyond two models to multiple capacity levels. Finally, dynamic mechanisms such as early exit or routing policies conditioned on the denoising state may further improve the speed--quality tradeoff.

We hope this work encourages more systematic study of timestep-dependent compute allocation for discrete diffusion language modeling and helps make masked diffusion LMs more efficient at inference time.

% \input{sections/sections_2}

% \newpage
\section*{Impact Statement}

This paper proposes an inference-time efficiency method for masked diffusion language models by scheduling denoising steps across models of different sizes. The primary intended impact is to reduce sampling computation, which can lower energy use, monetary cost, and associated carbon emissions from running and evaluating generative models. Improved efficiency can also broaden accessibility for researchers and practitioners with limited compute budgets, and may reduce concentration of advanced generative modeling work within a small number of well-resourced organizations.

These potential benefits have environmental and distributive dimensions. Large-scale model training and inference consume substantial electricity; lowering per-sample compute can reduce the footprint of experimentation and deployment and, depending on the energy mix, reduce greenhouse gas emissions. To the extent that compute and energy costs are passed on to communities through higher energy demand and pollution externalities, efficiency improvements can contribute (at the margin) to mitigating those burdens. However, the net environmental impact is ambiguous: efficiency gains can also lead to increased overall usage (a rebound effect), potentially offsetting per-sample savings if deployment scales up.

At the same time, making text generation cheaper and easier to deploy can amplify existing misuse risks associated with generative language models, including spam, phishing, misinformation, and other forms of automated manipulation, by increasing the feasible volume of generated content. This work does not introduce new model capabilities beyond what is already present in the underlying models; it primarily changes how computation is allocated during sampling. Nevertheless, we recommend that any deployment of models benefiting from these efficiency improvements follows standard responsible-use practices, such as access controls, abuse monitoring, rate limiting, and content safety filtering, and that evaluations consider both performance and environmental implications (e.g., reporting compute and energy metrics alongside quality).

% \bibliography{example_paper}
\bibliography{references/text-diffusion, references/manual_refs}
\bibliographystyle{icml2026}

%%%%%%%%%%%%%%%%%%%%%%%%%%%%%%%%%%%%%%%%%%%%%%%%%%%%%%%%%%%%%%%%%%%%%%%%%%%%%%%
%%%%%%%%%%%%%%%%%%%%%%%%%%%%%%%%%%%%%%%%%%%%%%%%%%%%%%%%%%%%%%%%%%%%%%%%%%%%%%%
% APPENDIX
%%%%%%%%%%%%%%%%%%%%%%%%%%%%%%%%%%%%%%%%%%%%%%%%%%%%%%%%%%%%%%%%%%%%%%%%%%%%%%%
%%%%%%%%%%%%%%%%%%%%%%%%%%%%%%%%%%%%%%%%%%%%%%%%%%%%%%%%%%%%%%%%%%%%%%%%%%%%%%%
\newpage
\appendix
% \onecolumn
\section{Additional Light Model Results}
\label{app:more_configs}

\begin{figure}
    \centering
    \includegraphics[width=1\linewidth]{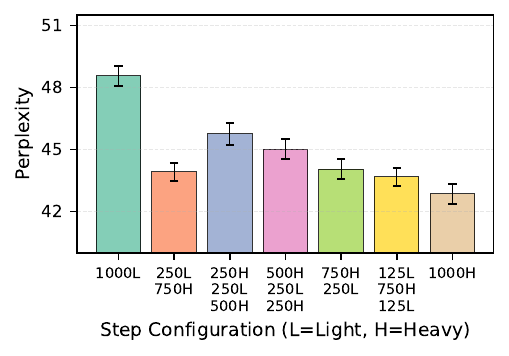}
    \caption{
    Generative perplexity (GPT-2) for hand-crafted model schedules using a heavy 12-block model and a light 6-block model with exactly 250/1000 light steps. Each bar label encodes a schedule as contiguous segments, e.g., $(\mathrm{L}125,\mathrm{H}750,\mathrm{L}125)$ denotes the \emph{sandwich} schedule (125 light steps, 750 heavy steps, 125 light steps).}
    \label{fig:12b-6b-barplot-perplexity}
\end{figure}

\begin{figure}
    \centering
    \includegraphics[width=1\linewidth]{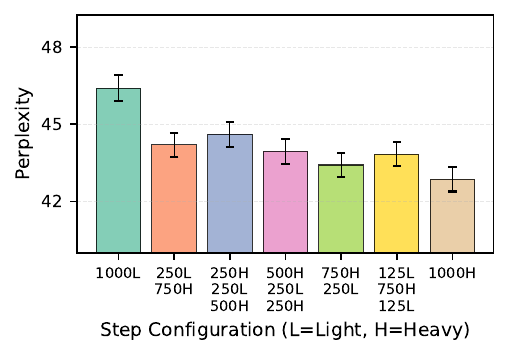}
    \caption{
    Generative perplexity (GPT-2) for hand-crafted model schedules using a heavy 12-block model and a light 8-block model with exactly 250/1000 light steps. Each bar label encodes a schedule as contiguous segments, e.g., $(\mathrm{L}125,\mathrm{H}750,\mathrm{L}125)$ denotes the \emph{sandwich} schedule (125 light steps, 750 heavy steps, 125 light steps).}
    \label{fig:12b-8b-barplot-perplexity}
\end{figure}

\begin{figure}
    \centering
    \includegraphics[width=1\linewidth]{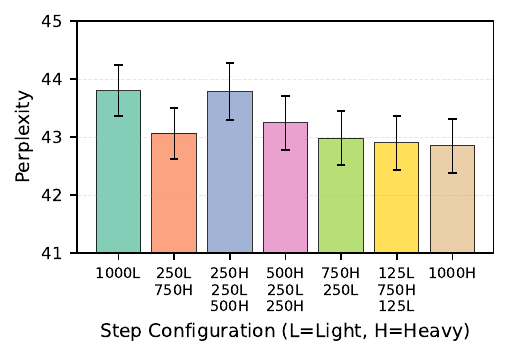}
    \caption{
    Generative perplexity (GPT-2) for hand-crafted model schedules using a heavy 12-block model and a light 10-block model with exactly 250/1000 light steps. Each bar label encodes a schedule as contiguous segments, e.g., $(\mathrm{L}125,\mathrm{H}750,\mathrm{L}125)$ denotes the \emph{sandwich} schedule (125 light steps, 750 heavy steps, 125 light steps).}
    \label{fig:12b-10b-barplot-perplexity}
\end{figure}

Additional hand-crafted schedules for light models with 6, 8, and 10 blocks are presented in Figures~\ref{fig:12b-6b-barplot-perplexity}, \ref{fig:12b-8b-barplot-perplexity}, and \ref{fig:12b-10b-barplot-perplexity} and exhibit the same qualitative pattern as the 4-block light model (Figure~\ref{fig:12b-4b-barplot-perplexity} in the main text).

\section{Additional Exhaustive Search Results}
\label{app:bottom20}

\begin{figure}[ht]
    \centering
    \includegraphics[width=1\linewidth]{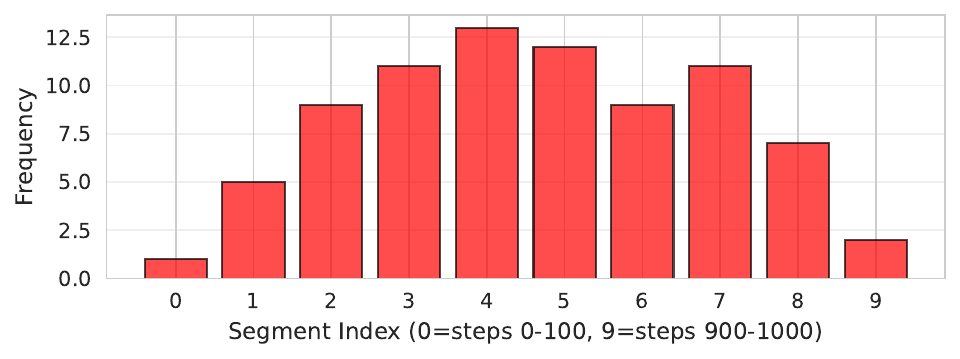}
    \caption{Segment frequency in the bottom 20 worst-performing configurations (highest perplexity) from the exhaustive search (Section~\ref{sec:bruteforce}). Bars show how often each segment is assigned to the light (4-block) model. Higher frequency suggests that replacing this segment is harmful.}
    \label{fig:bottom20config_barplot}
\end{figure}

Figure~\ref{fig:bottom20config_barplot} complements Figure~\ref{fig:top20config_barplot} in the main text by showing segment frequency among the 20 \emph{worst}-performing schedules from the exhaustive search (Section~\ref{sec:bruteforce}). Middle segments dominate, confirming that replacing them is most harmful.

\section{LM1B Generalization}
\label{app:lm1b}

\rev{To test whether our findings generalize beyond OpenWebText, we trained an identical model family on the One Billion Word Benchmark (LM1B)~\cite{chelba2014lm1b} with 128-token sequence length (see Section~\ref{sec:setup} for details).}

\begin{figure}
    \centering
    \includegraphics[width=1\linewidth]{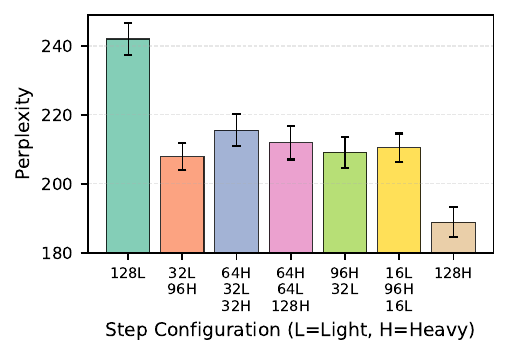}
    \caption{
    \rev{Generative perplexity for hand-crafted model schedules on LM1B (128-token context, 4-block light / 12-block heavy, 250/1000 light steps). The middle-step sensitivity pattern from Figure~\ref{fig:12b-4b-barplot-perplexity} reproduces on a different dataset and sequence length. Error bars correspond to 95\% confidence intervals.}}
    \label{fig:lm1b-12b-4b-barplot-perplexity}
\end{figure}

\rev{Figure~\ref{fig:lm1b-12b-4b-barplot-perplexity} and Table~\ref{tab:schedule_ppl_entropy_lm1b_12b-4b} show the hand-crafted schedule comparison on LM1B. The same qualitative pattern as on OpenWebText (Figure~\ref{fig:12b-4b-barplot-perplexity}) is observed: middle-step replacement yields the worst perplexity, while endpoint and sandwich placements perform best.}

\begin{table}[ht]
\centering
\begin{tabular}{@{}l r r@{}}
\toprule
Schedule & Gen.\ PPL {\scriptsize (CI)} & Entropy \\
\midrule
L128 & 242.0 {\scriptsize\,$\pm$\,4.69} & 4.30 \\
L32 $\to$ H96 & 207.9 {\scriptsize\,$\pm$\,3.96} & 4.25 \\
H32 $\to$ L32 $\to$ H64 & 215.5 {\scriptsize\,$\pm$\,4.66} & 4.24 \\
H64 $\to$ L32 $\to$ H32 & 211.9 {\scriptsize\,$\pm$\,4.83} & 4.24 \\
H96 $\to$ L32 & 209.1 {\scriptsize\,$\pm$\,4.48} & 4.24 \\
L16 $\to$ H96 $\to$ L16 & 210.5 {\scriptsize\,$\pm$\,4.16} & 4.26 \\
H128 & 188.8 {\scriptsize\,$\pm$\,4.35} & 4.21 \\
\bottomrule
\end{tabular}
\caption{\rev{Generative perplexity (with 95\% CI) and token-level entropy for hand-crafted model schedules on LM1B (128-token context, 4-block light / 12-block heavy, 32/128 light steps). \textbf{H}eavy: 12-block, \textbf{L}ight: 4-block. The same schedule ranking as on OpenWebText (Table~\ref{tab:schedule_ppl_entopy_owt_12b-4b}) is reproduced. Entropy variation across schedules is $<$0.06 nats.}}
\label{tab:schedule_ppl_entropy_lm1b_12b-4b}
\end{table}

\begin{figure}
    \centering
    \includegraphics[width=1\linewidth]{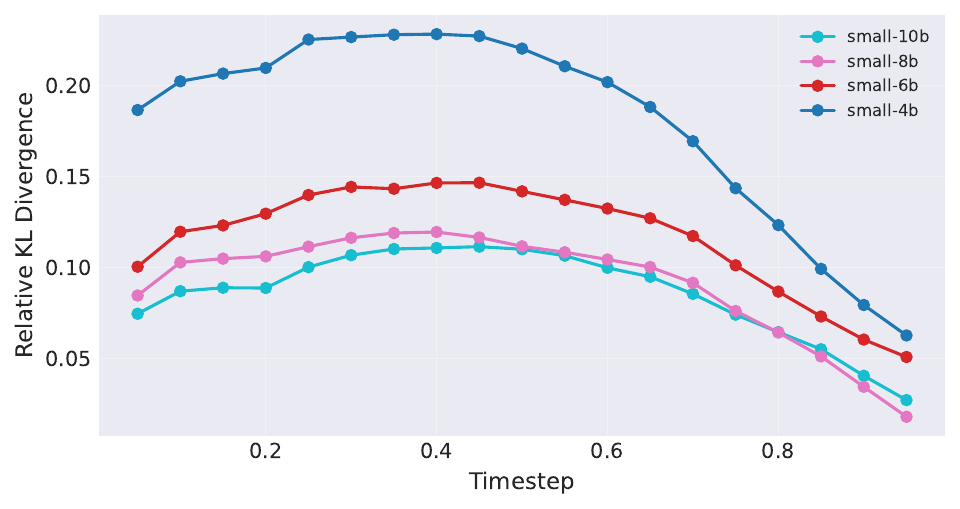}
    \caption{\rev{Relative token-level KL divergence (Eq.~\ref{eq:token_kl}) between model pairs trained on LM1B across timesteps, after subtracting a baseline KL curve computed between two independently trained heavy (12-block) checkpoints. The same middle-trajectory peak observed on OpenWebText (Figure~\ref{fig:kldiv-plot}) is reproduced, confirming cross-dataset generality.}}
    \label{fig:lm1b-kldiv-plot}
\end{figure}

\rev{We also replicate the KL-divergence analysis (Section~\ref{sec:similarity}) on the LM1B model family. Figure~\ref{fig:lm1b-kldiv-plot} shows the same characteristic peak in the middle of the denoising trajectory, confirming that the non-monotonic step-importance pattern is a general property of masked diffusion rather than an artifact of the OpenWebText setup.}

\section{Unconditional Generation Entropy}
\label{app:entropy}

\rev{Table~\ref{tab:schedule_ppl_entopy_owt_12b-4b} reports token-level entropy alongside generative perplexity for the unconditional OpenWebText schedules shown in Figure~\ref{fig:12b-4b-barplot-perplexity}. Entropy remains stable across all schedules (range 5.27--5.30), indicating that model scheduling does not reduce sample diversity even under substantial model substitution.}

\begin{table}[ht]
\centering
\begin{tabular}{@{}l r r@{}}
\toprule
Schedule & Gen.\ PPL {\scriptsize (CI)} & Entropy \\
\midrule
L1000 & 53.4 {\scriptsize\,$\pm$\,0.62} & 5.27 \\
L250 $\to$ H750 & 44.6 {\scriptsize\,$\pm$\,0.50} & 5.28 \\
H250 $\to$ L250 $\to$ H500 & 48.0 {\scriptsize\,$\pm$\,0.54} & 5.30 \\
H500 $\to$ L250 $\to$ H250 & 47.1 {\scriptsize\,$\pm$\,0.52} & 5.29 \\
H750 $\to$ L250 & 45.4 {\scriptsize\,$\pm$\,0.50} & 5.29 \\
L125 $\to$ H750 $\to$ L125 & 44.3 {\scriptsize\,$\pm$\,0.50} & 5.27 \\
H1000 & 42.9 {\scriptsize\,$\pm$\,0.47} & 5.29 \\
\bottomrule
\end{tabular}
\caption{\rev{Unconditional generative perplexity (with 95\% CI) and token-level entropy for OpenWebText schedules (same data as Figure~\ref{fig:12b-4b-barplot-perplexity}). \textbf{H}eavy: 12-block, \textbf{L}ight: 4-block. Entropy variation across schedules is $<$0.03 nats, confirming stable sample diversity.}}
\label{tab:schedule_ppl_entopy_owt_12b-4b}
\end{table}

\section{Prefix-Conditional Generation}
\label{app:prefix_cond}

\begin{figure}
    \centering
    \includegraphics[width=1\linewidth]{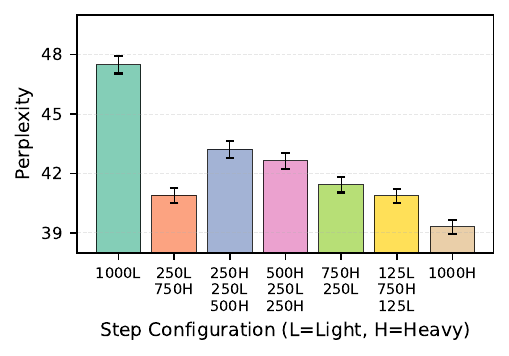}
    \caption{\rev{Conditional generative perplexity for hand-crafted schedules on OpenWebText with 256-token prefixes (4-block light / 12-block heavy, 250/1000 light steps). The schedule ranking matches the unconditional setting (Figure~\ref{fig:12b-4b-barplot-perplexity}). Error bars correspond to 95\% confidence intervals.}}
    \label{fig:owt_perplexity_barplot_12b4b_prefix256}
\end{figure}

\begin{figure}
    \centering
    \includegraphics[width=1\linewidth]{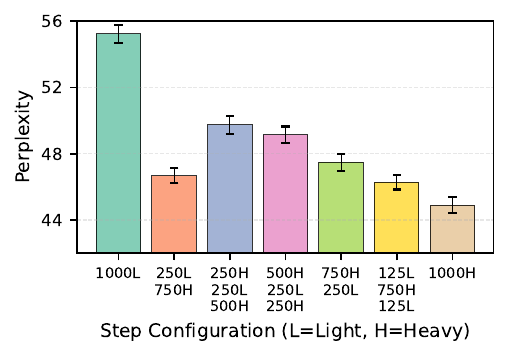}
    \caption{\rev{Conditional generative perplexity for hand-crafted schedules on OpenWebText with 128-token prefixes (4-block light / 12-block heavy, 250/1000 light steps). The pattern is consistent with the 256-token prefix setting (Figure~\ref{fig:owt_perplexity_barplot_12b4b_prefix256}). Error bars correspond to 95\% confidence intervals.}}
    \label{fig:owt_perplexity_barplot_12b4b_prefix128}
\end{figure}

\rev{To complement the prefix-conditional results with 256-token prefixes reported in the main text (Table~\ref{tab:owt_schedule_ppl_prefix256}), we provide the corresponding bar plot (Figure~\ref{fig:owt_perplexity_barplot_12b4b_prefix256}) and an additional evaluation with 128-token prefixes (Figure~\ref{fig:owt_perplexity_barplot_12b4b_prefix128}, Table~\ref{tab:owt_schedule_ppl_prefix128}). In both settings, the same schedule ranking holds: middle-step replacement is most harmful, while the sandwich schedule performs best among mixed schedules.}

\begin{table}[ht]
\centering
\begin{tabular}{@{}l r r@{}}
\toprule
Schedule & Cond.\ Gen.\ PPL {\scriptsize (CI)} & Entropy \\
\midrule
L1000 & 55.3 {\scriptsize\,$\pm$\,0.54} & 5.39 \\
L250 $\to$ H750 & 46.7 {\scriptsize\,$\pm$\,0.45} & 5.39 \\
H250 $\to$ L250 $\to$ H500 & 49.7 {\scriptsize\,$\pm$\,0.53} & 5.39 \\
H500 $\to$ L250 $\to$ H250 & 49.1 {\scriptsize\,$\pm$\,0.51} & 5.38 \\
H750 $\to$ L250 & 47.5 {\scriptsize\,$\pm$\,0.49} & 5.38 \\
L125 $\to$ H750 $\to$ L125 & 46.3 {\scriptsize\,$\pm$\,0.43} & 5.39 \\
H1000 & 44.9 {\scriptsize\,$\pm$\,0.46} & 5.38 \\
\bottomrule
\end{tabular}
\caption{\rev{Prefix-conditional generation (128-token prefix from OpenWebText). \textbf{H}eavy: 12-block, \textbf{L}ight: 4-block. The same schedule ranking as the 256-token prefix setting (Table~\ref{tab:owt_schedule_ppl_prefix256}) is observed. Entropy remains stable across schedules.}}
\label{tab:owt_schedule_ppl_prefix128}
\end{table}

%%%%%%%%%%%%%%%%%%%%%%%%%%%%%%%%%%%%%%%%%%%%%%%%%%%%%%%%%%%%%%%%%%%%%%%%%%%%%%%
%%%%%%%%%%%%%%%%%%%%%%%%%%%%%%%%%%%%%%%%%%%%%%%%%%%%%%%%%%%%%%%%%%%%%%%%%%%%%%%

\end{document}